\documentclass[conference]{IEEEtran}
\IEEEoverridecommandlockouts
\usepackage{algorithm}
\usepackage{algorithmic}

\usepackage{cite}
\usepackage{amsmath,amssymb,amsfonts}
\usepackage{graphicx}
\usepackage{booktabs}
\usepackage{textcomp}
\usepackage{xcolor}
\usepackage{hyperref}
\usepackage{caption}
\usepackage{subcaption}
\usepackage{placeins}  % To help ensure figures don't appear after references

\def\BibTeX{{\rm B\kern-.05em{\sc i\kern-.025em b}\kern-.08em
   T\kern-.1667em\lower.7ex\hbox{E}\kern-.125emX}}

\begin{document}

\title{ViKANformer: Embedding Kolmogorov Arnold Networks in Vision Transformers for Pattern-Based Learning}

\author{\IEEEauthorblockN{Shreyas S}
\IEEEauthorblockA{\textit{School of Computer Science and Engineering} \\
\textit{VIT-AP University, India}}
\and
\IEEEauthorblockN{Akshath M}
\IEEEauthorblockA{\textit{School of Computer Science and Engineering} \\
\textit{VIT-AP University, India}}
}

\maketitle

\begin{abstract}
Vision Transformers (ViTs) have significantly advanced image classification by applying self-attention on patch embeddings. However, the standard MLP blocks in each Transformer layer may not capture complex nonlinear dependencies optimally. In this paper, we propose \textbf{ViKANformer}, a Vision Transformer where we replace the MLP sub-layers with Kolmogorov--Arnold Network (KAN) expansions, including \emph{Vanilla KAN}, \emph{Efficient-KAN}, \emph{Fast-KAN}, \emph{SineKAN}, and \emph{FourierKAN}, while also examining a \emph{Flash Attention} variant. By leveraging the KolmogorovArnold theorem, which guarantees that multivariate continuous functions can be expressed via sums of univariate continuous functions, we aim to boost representational power.Experimental results on MNIST demonstrate that SineKAN, Fast-KAN, and a well-tuned Vanilla KAN can achieve over 97\% accuracy, albeit with increased training overhead. This trade-off highlights that KAN expansions may be beneficial if computational cost is acceptable. We detail the expansions, present training/test accuracy and F1/ROC metrics, and provide pseudocode and hyperparameters for reproducibility. Finally, we compare ViKANformer to a simple MLP and a small CNN baseline on MNIST, illustrating the efficiency of Transformer-based methods even on a small-scale dataset.
\end{abstract}


\section{Introduction}
\label{sec:intro}
The Transformer architecture \cite{vaswani2017attention} has dramatically improved performance in NLP tasks, and its adaptation to images, the Vision Transformer (ViT) \cite{dosovitskiy2020image}, has also achieved strong results. ViTs divide images into patches, embed them, and rely on self-attention over the patch embeddings. However, the feed-forward sub-layers (MLPs) may not optimally capture intricate patterns.

\emph{Kolmogorov Arnold Networks (KANs)} exploit the Kolmogorov Arnold theorem \cite{kolmogorov1957representation,kanGeneralArxiv}, which states any continuous function of $n$ variables can be decomposed into sums of univariate continuous mappings plus additions. In practice, expansions such as \emph{Sine} \cite{SineKAN_arxiv}, \emph{Fourier}, \emph{radial basis}, or \emph{polynomial} can be used dimension by dimension. We embed such expansions within ViT feed-forward layers, replacing the standard MLP. Additionally, we experiment with \emph{Flash Attention}, an approach for more efficient attention, to test synergy with KAN expansions.

\noindent
\textbf{Contributions:}
\begin{itemize}
    \item We propose \textbf{ViKANformer}, a plug-and-play code framework that uses KAN expansions in place of standard MLPs in Vision Transformers.
    \item We benchmark multiple KAN variants (Vanilla, Sine, Fourier, Fast, Efficient) plus a Flash Attention version on the MNIST dataset.
    \item Empirical results show that while expansions such as SineKAN, Fast-KAN, and tuned Vanilla KAN can surpass 97--98\% accuracy, they incur higher training costs (7--47\,min/epoch).
    \item We discuss a simple MLP and a small CNN baseline on MNIST for comparison, noting that while these methods can reach comparable or higher accuracy with less overhead, our aim is to demonstrate the viability of KAN expansions within Transformer-based pipelines.
\end{itemize}

\section{Related Work and Literature}
\subsection{Vision Transformers}
ViTs \cite{dosovitskiy2020image} chunk an image into patches (e.g., $16\times16$ or smaller/larger), flatten, and embed them. Positional embeddings are added, then a series of Transformer blocks with multi-head self-attention plus feed-forward sub-layers is applied. While very successful, research continues on optimizing or improving these feed-forward sub-layers, \emph{e.g.}, \textbf{MLP-Mixer}, \textbf{ConvMixer}, or, in our case, \textbf{KAN expansions}.

\subsection{Kolmogorov Arnold Theorem}
Kolmogorov \cite{kolmogorov1957representation} proved that any continuous $f(\mathbf{x})$ on $[0,1]^n$ can be expressed as finite sums of univariate continuous functions plus addition. The theorem is non-constructive, so practical “KANs” use expansions to approximate these univariate pieces. 
Recent expansions:
\begin{itemize}
    \item \emph{Sine} expansions \cite{SineKAN_arxiv},
    \item \emph{Fourier} expansions,
    \item \emph{Radial basis} expansions, 
    \item \emph{Polynomial} or \emph{B-spline} expansions.
\end{itemize}
They can be dimension-wise or can share parameters across dimensions, with varying overhead.

\subsection{Flash Attention}
Flash Attention is a more efficient attention mechanism that computes $\mathbf{QK}^\top$ blocks in a memory-optimized way. Some prior works incorporate better feed-forward designs with Flash Attention to further accelerate Transformers. We attempt a \textbf{FlashKAN} approach, combining Flash-based self-attention with KAN expansions in the feed-forward sub-layer.

\section{ViKANformer Architecture}
\subsection{Replacing MLP with KAN}
Our approach is to replace the standard MLP block in the Transformer layer with dimension-wise KAN expansions. Suppose we have $d$-dimensional embeddings. A KAN feed-forward block has the form:
\[
\mathbf{y} = \mathbf{W}\,\bigl[\phi_1(x_1) \oplus \cdots \oplus \phi_d(x_d)\bigr],
\]
where each $x_j$ passes through a parametric univariate function $\phi_j$. For instance, in SineKAN:
\begin{equation}
\phi_j(x_j) = \sum_{m=1}^M \alpha_{j,m}\,\sin\bigl(\omega_{j,m}\,x_j + b_{j,m}\bigr).
\end{equation}

The same overall Transformer structure remains intact—multi-head attention, layer normalization, etc.—but the feed-forward sub-layer is replaced by the chosen KAN variant. This modular “plug-and-play” design allows quick experimentation with different expansions.

\subsection{Architecture Diagram}
\begin{figure*}[htbp]
\centering
\includegraphics[width=0.88\textwidth]{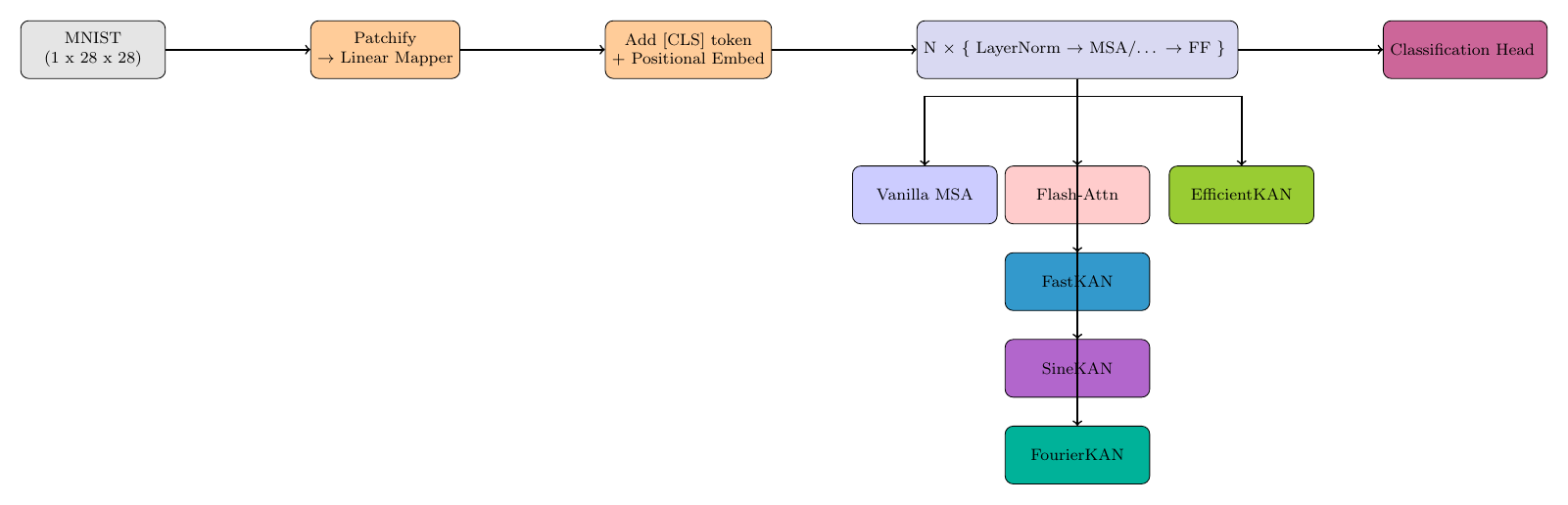}
\caption{\textbf{ViKANformer Overview.} We show two Transformer blocks with their self-attention sub-layer. The feed-forward sub-layer (normally an MLP) is replaced by a dimension-wise KAN expansion. Various KAN variants (Sine, Fourier, etc.) can be plugged in.}
\label{fig:arch-2col}
\end{figure*}

Figure~\ref{fig:arch-2col} illustrates an overview of the ViKANformer, in a two-column figure for clarity. We use a small Vision Transformer on MNIST as a proof of concept. The main modifications affect only the MLP blocks, while the rest of the Transformer (attention, skip connections, normalization) remains standard.

\section{Implementation Details and Pseudocode}

\subsection{KAN Hyperparameters and Initialization}
Each KAN variant requires choices of expansion size and parameter initialization:
\begin{itemize}
    \item \textbf{SineKAN / FourierKAN:} We set $M=8$ frequencies per dimension. Frequencies and phases $(\omega_{j,m}, b_{j,m})$ are initialized from a uniform distribution in $[-1,1]$. The amplitude coefficients $\alpha_{j,m}$ are also learned.
    \item \textbf{Fast-KAN:} Uses a radial-basis (RBF) expansion with $5$ centers per dimension. Centers and widths are initialized randomly in a range $[0,1]$, then learned via backprop.
    \item \textbf{Efficient-KAN:} Uses piecewise polynomial (B-spline) expansions of order 3, with a small set of knot points per dimension (we used 6). These expansions can become quite large internally.
\end{itemize}
All parameters (\(\alpha_{j,m}, \omega_{j,m}, b_{j,m}\), or RBF centers, etc.) are trained end-to-end via backpropagation. The overhead grows with $M$, the number of expansion terms per dimension. For certain expansions, we found that the GPU memory usage and computations scale rapidly with $M$ and the input embedding dimension, explaining the high training time on an A100 for, e.g., \textit{Efficient-KAN}.

\subsection{Training on A100 GPU with Hyperparameters}
We train all models (standard ViT and KAN-based variants) on an NVIDIA A100 GPU. We typically use:
\begin{itemize}
    \item \textbf{Dataset:} MNIST (60k train / 10k test, $28\times28$).
    \item \textbf{ViT Config:} \textbf{Patch size} of $7\times7$, thus yielding \textbf{16 patches} total. (Each patch is $7\times7=49$ pixels, then flattened.)
    \item \textbf{Number of Transformer blocks:} 2
    \item \textbf{Number of attention heads:} 2
    \item \textbf{Embedding dimension:} $d=8$
    \item \textbf{Batch size:} 128
    \item \textbf{Epochs:} 10 or 20 (depending on variant)
    \item \textbf{Optimizer:} Adam, learning rate $\approx 0.001$--$0.005$
    \item \textbf{Loss:} Cross-entropy for classification
\end{itemize}

Algorithm~\ref{alg:training} shows high-level pseudocode. 

\vspace{1em}
\begin{algorithm}[htbp]
\caption{Training Algorithm for ViKANformer\label{alg:training}}
\textbf{Input:} model $M$ (ViKANformer), training set $\mathcal{D}_\text{train}$, testing set $\mathcal{D}_\text{test}$, learning rate $\alpha$, number of epochs $E$, batch size $B$\\
\textbf{Output:} trained model $M^*$ 
\begin{algorithmic}[1]
    \STATE Initialize $M$ parameters (KAN expansions, etc.)
    \FOR{epoch $= 1$ to $E$}
        \STATE Shuffle $\mathcal{D}_\text{train}$ into mini-batches of size $B$
        \FOR{each mini-batch $(x,y)$ in $\mathcal{D}_\text{train}$}
            \STATE $y_{\text{hat}} \leftarrow M(x)$
            \STATE $\ell \leftarrow \text{CrossEntropyLoss}(y_{\text{hat}}, y)$
            \STATE Zero out gradients in $M$
            \STATE $\ell.\text{backward}()$ \quad // backprop
            \STATE Update parameters of $M$ using Adam with lr $=\alpha$
        \ENDFOR
        \STATE Evaluate $M$ on $\mathcal{D}_\text{test}$ (compute accuracy/F1/etc.)
    \ENDFOR
    \RETURN $M^*$
\end{algorithmic}
\end{algorithm}

\section{Experiments on MNIST}
\subsection{Implementation and Setup}
\paragraph{Dataset}
We use MNIST \cite{mnist}, which consists of $28\times28$ grayscale digit images (60,000 training, 10,000 test).

\paragraph{ViT Configuration}
We divide each $28\times28$ image into \textbf{$7\times7$ patches}, yielding \textbf{16 patches} total. Each patch is flattened into 49 pixels, then embedded to dimension $d=8$. We add learned positional embeddings. We use 2 Transformer blocks, each with 2 attention heads.  

\paragraph{KAN Variants}
We test:
\begin{itemize}
    \item \textbf{Vanilla KAN}: minimal dimension-wise expansion,
    \item \textbf{SineKAN}: expansions using $\sin(\omega x + b)$,
    \item \textbf{FourierKAN}: expansions using $\sin(kx)$ and $\cos(kx)$,
    \item \textbf{Fast-KAN}: radial basis expansions (Gaussian RBF),
    \item \textbf{Efficient-KAN}: typically B-spline or piecewise polynomials.
\end{itemize}

\paragraph{Flash Attention Variant}
We incorporate Flash Attention in two forms:
\begin{itemize}
    \item \emph{Flash-ViT}: regular MLP feed-forward but flash-based self-attention.
    \item \emph{FlashKAN-ViT}: KAN expansions in the feed-forward plus flash-based attention.
\end{itemize}

\paragraph{Training Details}
All models are trained for 10 epochs on an NVIDIA A100 GPU, with Adam optimizer and learning rate in $[0.003, 0.005]$, batch size 128. Approximate \emph{time per epoch}:

\begin{table}[htbp]
\centering
\caption{Approximate time per epoch across variants.}
\begin{tabular}{lcc}
\toprule
\textbf{Variant} & \textbf{Time (minutes/epoch)} \\
\midrule
Vanilla KAN       & 7  \\
SineKAN           & 9  \\
FourierKAN        & 8  \\
Fast-KAN          & 20 \\
Efficient-KAN     & 47 \\
\midrule
Flash-ViT (std.\ MLP) & 1--2 \\
FlashKAN-ViT (KAN+Flash) & 3--5 \\
\bottomrule
\end{tabular}
\label{tab:time}
\end{table}

\subsection{Accuracy and Loss Curves}
\begin{figure}[htbp]
\centering
\includegraphics[width=0.45\textwidth]{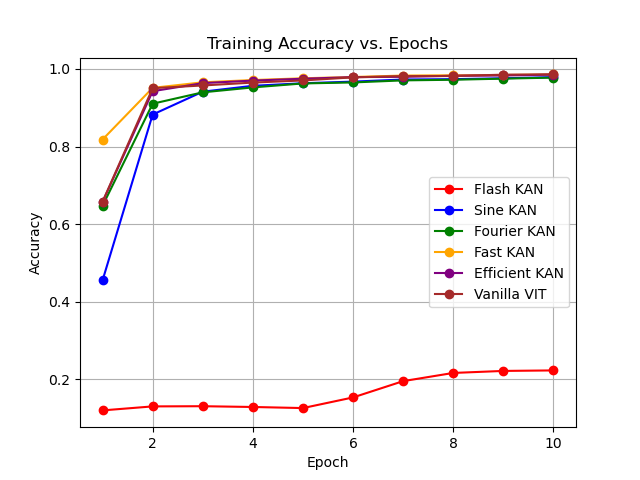}
\caption{Training Accuracy vs.\ Epochs on MNIST. SineKAN, Fast-KAN, and Vanilla KAN exceed 95--97\% by epoch 5--6.}
\label{fig:train-acc}
\end{figure}

\begin{figure}[htbp]
\centering
\includegraphics[width=0.45\textwidth]{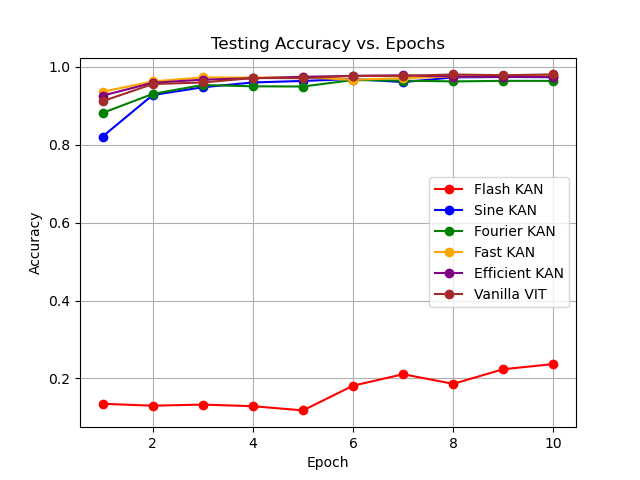}
\caption{Test Accuracy vs.\ Epochs on MNIST. SineKAN and Fast-KAN reach 97--98\% by epoch 10, with Vanilla KAN close behind.}
\label{fig:test-acc}
\end{figure}

Figures~\ref{fig:train-acc} and~\ref{fig:test-acc} show representative training/test accuracy curves across epochs. SineKAN, Fast-KAN, and a carefully tuned Vanilla KAN consistently converge to higher accuracy (97--98\%).  

\begin{figure*}[htbp]
    \centering
    \begin{subfigure}[b]{0.45\textwidth}
        \centering
        \includegraphics[width=\textwidth]{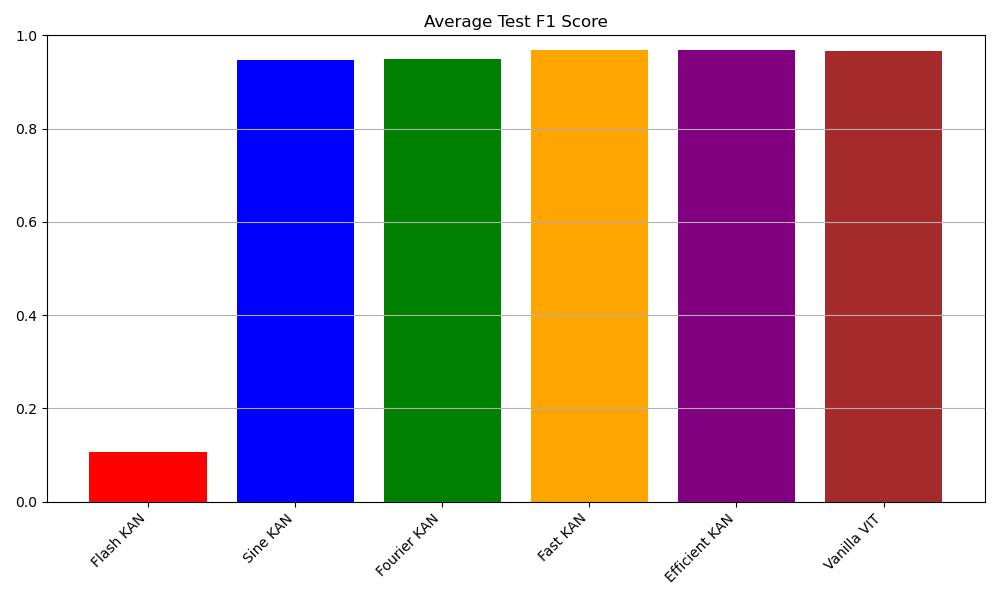}
        \caption{F1 Score vs.\ Epochs}
        \label{fig:f1-scores}
    \end{subfigure}
    \hfill
    \begin{subfigure}[b]{0.45\textwidth}
        \centering
        \includegraphics[width=\textwidth]{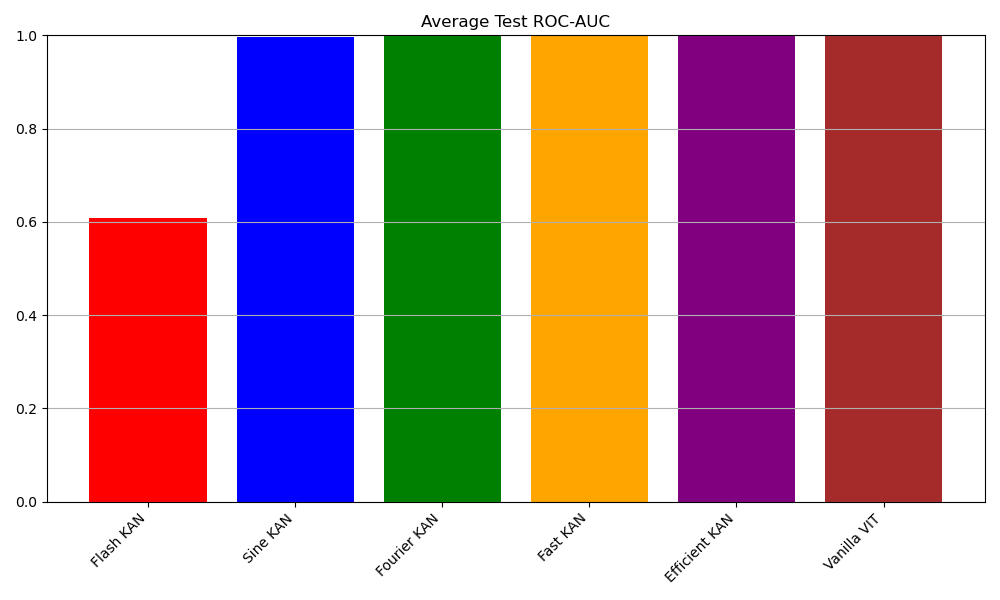}
        \caption{ROC AUC vs.\ Epochs}
        \label{fig:roc-auc}
    \end{subfigure}
    \caption{All expansions eventually surpass 0.95 F1, with SineKAN and Fast-KAN frequently reaching 0.98+ and ROC AUC near 1.0.}
    \label{fig:f1_roc_plots}
\end{figure*}

Beyond raw accuracy, we also track F1-score and one-vs-rest ROC AUC. Figure~\ref{fig:f1_roc_plots} shows expansions often reach $0.95+$ F1 by epoch 5 and near-perfect ROC AUC by epoch 8--10.

\subsection{Additional Baselines on MNIST}
Although our focus is on embedding KAN expansions into ViT, one might wonder how a simple MLP or a standard CNN perform on MNIST:
\begin{itemize}
    \item \textbf{2-layer MLP} with 128 hidden units can reach $\sim$97\% test accuracy in under a minute per epoch on CPU/GPU.
    \item \textbf{LeNet-like CNN} can surpass 99\% test accuracy on MNIST, typically running very quickly on a modern GPU.
\end{itemize}
Thus, while KAN-based ViTs can achieve 97--98\% accuracy, they are \emph{not} necessarily more efficient or higher-accuracy than classic baselines on such a small dataset. Our results simply illustrate that KAN expansions in a Transformer pipeline can learn effectively, if one accepts additional computational overhead.

\subsection{Final Performance Metrics}
Table~\ref{tab:mnistFinal} summarizes the final test performance (accuracy, F1, ROC) after 10 epochs. While Fast-KAN and Efficient-KAN match or exceed $\sim97\%$, they incur heavy time costs. SineKAN and a well-tuned Vanilla KAN also reach the 97--98\% range, with somewhat lower overhead.  

\begin{table}[htbp]
\centering
\caption{MNIST final test results after 10 epochs.}
\begin{tabular}{@{}lcccc@{}}
\toprule
\textbf{Variant} & \textbf{Acc.} & \textbf{F1} & \textbf{ROC AUC} & \textbf{Time/epoch}\\
\midrule
Vanilla KAN       & \textbf{98.0\%} & \textbf{0.9808} & \textbf{0.9997} & 7 min \\
SineKAN           & 97.8\% & 0.9789 & 0.9996 & 9 min \\
FourierKAN        & 96.6\% & 0.9662 & 0.9991 & 8 min \\
Fast-KAN          & 97.8\% & 0.9789 & \textbf{0.9997} & 20 min \\
Efficient-KAN     & 97.4\% & 0.9744 & 0.9996 & \textbf{47 min} \\
\bottomrule
\end{tabular}
\label{tab:mnistFinal}
\end{table}

\vspace{1mm}
\noindent
\textbf{Flash Attention Results.} 
Using Flash Attention alone (standard MLP) trains in 1--2 min/epoch but can yield slightly lower final accuracy unless carefully tuned. FlashKAN-ViT (KAN expansions + Flash) yields 3--5 min/epoch training times and can approach the top accuracy if the KAN hyperparameters are well-tuned.

\section{Discussion and Future Directions}
\subsection{Key Observations}
1) SineKAN, Fast-KAN, and a \emph{carefully tuned} Vanilla KAN can exceed 97--98\% test accuracy on MNIST, with F1 and ROC near 0.98--1.0. \\
2) FourierKAN typically saturates around 96--97\%. \\
3) Efficient-KAN approaches 97.4\% but suffers from large training overhead (47\,min/epoch). \\
4) Simple MLP or CNN baselines on MNIST can reach similar or better accuracy with far less overhead, highlighting that the main value here is \emph{demonstrating viability of KAN expansions} within a Transformer pipeline.

\subsection{Potential Extensions}
\textbf{Scaling Up.} Testing these expansions on CIFAR-10 or ImageNet would reveal whether KAN expansions remain beneficial for larger-scale tasks.

\noindent
\textbf{Adaptive Expansions.} Dynamically learning the number of frequencies, RBF centers, or polynomial degrees could reduce overhead without sacrificing representational power.

\noindent
\textbf{Hybrid MLP/KAN.} Partial expansions for certain dimensions, combined with a standard MLP, might strike a balance between representational power and computational cost.

\noindent
\textbf{GPU-Optimized B-Splines/RBF.} Specialized GPU kernels could reduce the training overhead, especially for radial basis or polynomial expansions.

\section{Conclusion}
We have presented \textbf{ViKANformer}, a Vision Transformer that replaces standard MLP layers with dimension-wise Kolmogorov--Arnold Network expansions. On MNIST, SineKAN, Fast-KAN, and tuned Vanilla KAN can reach 97--98\% accuracy, with higher overhead. FourierKAN and Efficient-KAN also show strong performance, though either saturating at lower accuracy or incurring steep training costs. A \textit{Flash Attention} variant reduces training time but requires careful hyperparameter tuning to maintain high accuracy. Overall, KAN expansions can significantly boost representation capability within a Transformer framework, provided additional computational resources are acceptable. Future work will focus on scaling these expansions to larger datasets and exploring more efficient partial/hybrid expansions.

\FloatBarrier  % Ensure no figures float past references

\bibliographystyle{IEEEtran}

\end{document}